%% file: main-arxiv.tex
\documentclass{article}

\usepackage{arxiv}

\usepackage[utf8]{inputenc}
\usepackage[T1]{fontenc}
\usepackage[english]{babel}
\usepackage{csquotes}
\usepackage[final,
            tracking=true,
            kerning=true]{microtype}
\usepackage{xspace}

\usepackage{amssymb}
\usepackage{amsmath}
\usepackage{mathtools}
\usepackage{bbm}
\let\leq\leqslant

\usepackage{graphicx}
\usepackage{booktabs}
\usepackage{adjustbox}

\usepackage{nth}
\usepackage{siunitx}
\usepackage{etoolbox}
\robustify\bfseries

\usepackage{enumitem}

\usepackage[dvipsnames]{xcolor}

\newcommand*{\titletext}{Interpretable Locally Adaptive Nearest Neighbors}

\usepackage[backend=biber,
            sorting=none,
            maxcitenames=2,
            maxbibnames=4,
            backref=true]{biblatex}
\usepackage{varioref}
\usepackage[hypertexnames=false,
            colorlinks=true,
            linkcolor=MidnightBlue,
            urlcolor=Maroon,
            citecolor=ForestGreen,
            pdfauthor={Jan Philip Göpfert},
            pdftitle={\titletext}]{hyperref}
\usepackage[all]{hypcap}
\usepackage[noabbrev,
            capitalize,
            nameinlink]{cleveref}

\usepackage[backgroundcolor=white,
            linecolor=lightgray,
            bordercolor=white,
            textsize=tiny]{todonotes}

\title{\titletext}

\usepackage[auth-sc]{authblk}

\author[1]{Jan Philip Göpfert}
\affil[1]{Bielefeld University, Germany}
\author[2]{Heiko Wersing}
\affil[2]{Honda Research Institute Europe GmbH, Offenbach, Germany}
\author[1]{Barbara Hammer}

\newcommand*{\ie}{i.\,e.\@\xspace}

\newcommand*{\etc}{etc.\@\xspace}

\newcommand*{\wrt}{w.\,r.\,t.\@\xspace}
\newcommand*{\psd}{p.\,s.\,d.\@\xspace}
\newcommand*{\R}{\mathbb{R}}

\newcommand*{\transpsymb}{\mathsf{T}}
\DeclarePairedDelimiterX{\norm}[1]{\lVert}{\rVert}{#1}

\newcommand*{\transmat}{\Omega}
\newcommand*{\quadrmat}{\Lambda}

\begin{document}
\maketitle
\begin{abstract}
\input{abstract.tex}
\end{abstract}
\input{content.tex}
\input{acknowledgements.tex}
\AtNextBibliography{\raggedright}
\printbibliography
\end{document}

%% file: abstract.tex
When training automated systems, it has been shown to be beneficial to adapt the
representation of data by learning a problem-specific metric. This metric is
global. We extend this idea and, for the widely used family of k nearest
neighbors algorithms, develop a method that allows learning \emph{locally}
adaptive metrics. These local metrics not only improve performance, but are
naturally interpretable. To demonstrate important aspects of how our approach
works, we conduct a number of experiments on synthetic data sets, and we show
its usefulness on real-world benchmark data sets.

%% file: content.tex
\section{Introduction}
Machine learning models increasingly pervade our daily lives in the form of
recommendation systems, computer vision, driver assistance, \etc, challenging us
to realize seamless cooperation between human and algorithmic agents. One
desirable property of predictions made by machine learning models is their
transparency, expressed in such a way as a statement about which factors of a
given setting have the greatest influence on the decision at hand -- in
particular, this requirement aligns with the EU General Data Protection
Regulations, which include a “right to explanation”~\autocite{eu-explain}. The
native transparency of machine learning models varies considerably based on the
form and complexity of the models, ranging from intuitive prototype-based
classifiers, which allow a substantiation of a decision in the form of a typical
class representative~\autocite{Kohonen1997SelfOrganizingMaps}, to mostly opaque
black-box models found in deep learning, for which additional posterior
explanation technologies are required~\autocite{inn}. Interestingly, several
popular interpretation technologies for black-box models rely on local feature
weighting schemes~\autocite{lime}. Moreover, machine learning models that are
intrinsically based on a feature relevance weighting~\autocite{Hammer2002GRLVQ},
enjoy a wide popularity in particular in medical domains to uncover relevant
insight, such as the discovery of potential biomarkers~\autocite{trisomie}.

Intuitive indications of which features are most or least relevant for a given
model's decision can be provided by metric-learning approaches, such as
GRLVQ~\autocite{Hammer2002GRLVQ}, which adapts a diagonal matrix, scaling the
relevance of the input features. Generalizations that use a full matrix, such as
GMLVQ~\autocite{Schneider2009AdaptiveRelevanceMatrices}, exist, but a
\emph{single global} quadratic matrix remains the most common
choice~\autocite{bellet}. Large margin nearest neighbor learning (LMNN)
implements this idea for a k-nearest neighbor (kNN) classification scheme
\autocite{Weinberger2009LargeMarginNearestNeighbors}.
A few approaches extend this setting to non-global matrices,
such as LGRLVQ~\autocite{Hammer2005LGRLVQ} and LGMLVQ~\autocite{Schneider2009AdaptiveRelevanceMatrices},
which can be accompanied by learning-theoretical guarantees,
but they allow only one matrix per prototype,
which corresponds to one metric per Voronoi cell in
the input space.
An extension of LMNN~\autocite{Weinberger2008FastDistanceMetric} requires an explicit partitioning of the training data and learns one metric per subset.
The partitioning, commonly based on the respective class labels,
is set before training and remains unchanged,
which makes the extension straight-forward but inflexible.
Parametric Local Metric Learning (PLML)~\autocite{Wang2012ParametricLocalMetric} learns a smooth metric matrix function over the data manifold,
but again, its specific metric matrices are based on so-called anchor points,
such as the means of clusters according to some supervised algorithm.
\Textcite{Noh2010GenerativeLocalMetric} take a different approach with Generative Local Metric Learning (GLML),
where they learn an optimal local metric for a learned generative model.
Fitting class-wise Gaussians, they inherit the inflexibilties that come with this assumption-heavy approach
and fail to learn well-performing metrics~\autocite{Wang2012ParametricLocalMetric}.
GLML does show promise for the special case where the number of training samples is further constrained.
Its original analysis is therefore focused on that particular setting.

In this work, we formulate and explore an extension of kNN to local relevance
matrices, which are \emph{specific to a given point} and indicate the
\emph{local relevance} of the features in its region, \ie the factors most
relevant for a \emph{specific decision} rather than the global model. Further,
unlike LMNN, PLML, and GLML, we implement an online adaptation technique, which
can be integrated into incremental models or models for streaming data, such as
the one proposed by \textcite{samknn}. In the following, we will propose a cost
function based on a differentiable approximation of the output label
distribution of a kNN classifier, and we will demonstrate how to derive an
intuitive local relevance learning scheme based thereon. To investigate the
resulting learned, local feature relevances, and to demonstrate how they aid in
interpreting data, we compute according low-dimensional embeddings.

This extension of the Locally Adaptive Nearest Neighbors~\autocite{Gopfert2020LocallyAdaptiveNearestNeighbors} contains an expanded evaluation with further datasets and an embedding-based demonstration of the usefulness of the learned metrics.

\section{Local metric learning for kNN classifiers}\label{section:lann}
Assume data $X=\{\vec x^1,\ldots,\vec x^m\}\subset\mathbb{R}^n$ are given, with
label $y_i$ for data point $\vec x^i$, where labels are element of a finite
number of $L$ different labels. Assume a number $k>0$ is fixed. A kNN classifier
crucially depends on a distance measure
$d:\mathbb{R}^n\times\mathbb{R}^n\to\mathbb{R}$. Given a data point $\vec x\in
\mathbb{R}^n$, the \emph{neighborhood} $N(\vec x)$ of $\vec x$ in $X$ is defined
as the set of $k$ points $\vec x^i$ in $X$ where $d(\vec x^i, \vec x)$ is
smallest. A weighted kNN classifier computes the \emph{support} $S$ for label
$y$ given input $\vec x$
\begin{equation*}
    S(y \mid \vec x)
  = \mspace{-15mu} \sum_{\substack{\vec x^i \in N(\vec x) \\ y_i = y}}
    \mspace{-9mu} \frac{1}{d(\vec x^i, \vec x)}
\end{equation*}
and outputs the label $y$ with maximum support. This definition relies on a
global distance measure $d$ such as the squared Euclidean distance measure
$d(\vec x^i,\vec x)=(\vec x^i-\vec x)^{\transpsymb}(\vec x^i-\vec x)$. Metric
learning such as LMNN~\autocite{Weinberger2009LargeMarginNearestNeighbors}
substitutes the Euclidean distance by a parameterized quadratic form
\begin{equation*}
      d_{\quadrmat}(\vec x^i,\vec x)
    = (\vec x^i-\vec x)^{\transpsymb} \quadrmat \mspace{3mu} (\vec x^i-\vec x)
\end{equation*}
with positive semi-definite (\psd) matrix $\quadrmat$, which is determined based
on given data. LMNN relies on the objective to change the distance such that
intruders, i.e.\ points $\vec x^i$ in $N(\vec x)$ which do not have the same
label as $\vec x$, are moved outside $N(\vec x)$ with a margin. This problem can
be phrased as a semi-convex constraint optimization problem for the metric
parameters $\quadrmat$~\autocite{Weinberger2009LargeMarginNearestNeighbors}.
LMNN uses a global distance measure, which does not necessarily resemble the
relevance of input features for the local decision $f(\vec x)$.

In the following, we want to ask and answer, whether it is possible to (i) learn
local metrics without a fixed prior decomposition of the space, and (ii) develop
an online learning scheme, which carries the potential of an integration into
streaming and incremental scenarios such as the self-adjusting-memory
kNN~\cite{samknn}. We assume a local distance measure
\begin{equation*}
    d_{\quadrmat_i}(\vec x^i,\vec x)
  = (\vec x^i-\vec x)^{\transpsymb} \quadrmat_i \mspace{2mu} (\vec x^i-\vec x)
\end{equation*}
where $d_{\quadrmat_i}$ is attached to the data point $\vec x^i$ and it is used
whenever the distance measure from $\vec x^i$ to another data point is computed.
Here, $\quadrmat_i$ is an adaptive \psd matrix, which can be parameterized as
$\quadrmat_i = (\transmat^i) (\transmat^i)^{\transpsymb}$ with possibly low-rank
matrix $\transmat^i\in \mathbb{R}^{n\times n'}$ for some $n'\leq n$ or even
diagonal form $\quadrmat_i = \mathrm{diag}\big((\lambda_1^i)^2, \dots,
(\lambda_n^i)^2\big)$.

Given an input $\vec x$ with desired output $y$, we can derive a stochastic
gradient scheme to adapt these metric parameters online as follows: We
approximate the output of a weighted kNN using the \emph{softmax} function with
parameter $\beta>0$, which yields a probability distribution over all possible
output labels $1, \dots, L$:
\begin{equation*}
    P(y \mid \vec x)
  := \left(
     \frac{\exp(S(y \mid \vec x) / \beta)}{\sum_{y'} \exp(S(y' \mid \vec x) / \beta)}
     \right)_{y=1, \dots, L} \mspace{-60mu} \in [0, 1]^L
\end{equation*}
where local metrics $d_{\quadrmat_i}$ are used to evaluate the support $S(y \mid
\vec x)$, which indicates the vector of probabilities of the $L$ output labels.
Assume a desired output $y=l$ is given, this induces a probability distribution
over the labels by its one-hot encoding in $\{0,1\}^L$, which we denote by $P(y
\mid l)$.

Then, a suitable loss function is offered by the Kullback-Leibler divergence,
resulting in the overall error
\begin{align*}
     E
  &= \sum_{i = 1}^m E(\vec x^i, y_i)
   = \sum_{i = 1}^m
     \mathrm{KL} \big(P(y \mid y_i) \;\big\|\; P(y \mid \vec x^i) \big) \\
  &= -\sum_{i = 1}^m \sum_{l = 1}^L
     P(y = l \mid y_i)
     \cdot
     \log \frac{P(y = l \mid \vec x^i)}{P(y = l \mid y_i)} \\
  &= -\sum_{i = 1}^m\log P(y = y_i \mid \vec x^i)
   = -\sum_{i = 1}^m
     \log
     \left(
       \frac{\exp(S(y_i \mid \vec x^i) / \beta)}
       {\sum_{y'} \exp(S(y' \mid \vec x^i) / \beta)}
     \right) \\
\end{align*}
since $P(y = l \mid y_i) = \delta_{l, y_i}$ (the Kronecker delta), where we
use the identity $0 \cdot \log 0 = 0$. For stochastic gradient descent, we
consider the derivative of a term \wrt metric parameters $\transmat^j_{kl}$
for a matrix $\quadrmat_j = (\transmat^j)(\transmat^j)^{\transpsymb}$. This
yields
\begin{align*}
   &\quad \frac{\partial}{\partial \transmat^j_{kl}}
    \Bigg(
      - \log \big(\exp(S(y_i \mid \vec x^i) / \beta) \big)
      + \log
      \bigg(
        {\sum_{y'} \exp(S(y' \mid \vec x^i) / \beta)}
      \bigg)
    \Bigg) \\
  &= -\frac{1}{\beta} \cdot \frac{\partial S(y_i \mid \vec x^i)}{\partial \transmat^j_{kl}} \\
    &\quad + \frac{1}{\beta} \cdot \frac{1}{\sum_{y'} \exp(S(y' \mid \vec x^i) / \beta)}
    \cdot \sum_{y'}
    \bigg(
      \frac{1}{\beta} \cdot \exp(S(y' \mid \vec x^i) / \beta) \cdot \frac{\partial S(y' \mid \vec x^i)}{\partial \transmat^j_{kl}}
    \bigg)
\end{align*}
Further,
\begin{align*}
     \frac{\partial S(y' \mid \vec x^i)}{\partial \transmat^j_{kl}}
  &= \frac{\partial}{\partial \transmat^j_{kl}}
     \sum_{\vec x^o\in N(\vec x^i), y_o=y'}\frac{1}{d_{\quadrmat_o}(\vec x^o,\vec x^i)}\\
  &=
     \begin{cases}
       -{1} / {d_{\quadrmat_j}(\vec x^j, \vec x^i)^2}
       \cdot \frac{\partial d_{\quadrmat_j}(\vec x^j, \vec x^i)}{\partial \transmat^j_{kl}}
       & \text{if } \vec x^j \in N(\vec x^i), y_j=y' \\
       0
       & \text{otherwise}
     \end{cases}
\end{align*}
yields the derivative $0$ for all $\quadrmat_j$ where $\vec x^j \not\in N(\vec
x^i)$. For neighbors $\vec x^j \in N(\vec x^i)$ we obtain
\begin{equation*}
  \frac{\partial E(\vec x^i, y_i)}{\partial \transmat^j_{kl}}
  = \begin{cases}
      \frac{1}{\beta \cdot d_{\quadrmat_j}(\vec x^j, \vec x^i)^2}
      \cdot
      \left(
        1 -
        \frac{\exp(S(y_i \mid \vec x^i) / \beta)}
        {\sum_{y'} \exp(S(y' \mid \vec x^i) / \beta)}
      \right)
      \cdot
      \frac{\partial d_{\quadrmat_j}(\vec x^j, \vec x^i)}
      {\partial \transmat^j_{kl}}
      & \text{if } y_j = y_i \\
      -\frac{1}{\beta \cdot d_{\quadrmat_j}(\vec x^j, \vec x^i)^2}
      \cdot
      \left(
      \frac{\exp(S(y_j \mid \vec x^i) / \beta)}
      {\sum_{y'} \exp(S(y' \mid \vec x^i) / \beta)}
      \right)
      \cdot
      \frac{\partial d_{\quadrmat_j}(\vec x^j, \vec x^i)}
      {\partial \transmat^j_{kl}}
      & \text{if } y_j \neq y_i
    \end{cases}
\end{equation*}
It is necessary to add a regularization step to prevent divergence of the
parameters, e.g.\ a soft or hard constraint for $\det \quadrmat_j$ or a
restriction of the norm of the diagonal of the matrices. If we chose the metrics
in the form of diagonal matrices $\quadrmat = \mathrm{diag}(\lambda_1^2, \dots,
\lambda_n^2)$, the derivative yields $\partial d_{\quadrmat}(\vec x, \vec x') /
\partial \lambda_l = 2 \lambda_l \cdot (x_l - x'_l)^2$. In this case, a
stochastic gradient descent directly corresponds to a Hebbian scheme: for $y_j =
y_i$, diagonal terms for those dimensions $l$ are enhanced (after normalization)
which correspond to small values $(x^j_l - x^i_l)^2$; for $y_j \neq y_i$, we
find the opposite. This behavior resembles popular metric learning schemes as
proposed in the context of prototype-based
classifiers~\autocite{Hammer2005LGRLVQ,Schneider2009AdaptiveRelevanceMatrices}.
Yet, while these technologies restrict metric forms to receptive fields of
prototypes, we are able to learn an individual weighting scheme for every data
point of the kNN classifier. Apart from the different objective, this fact -- a
local weighting scheme -- is the most distinguishing feature of the proposed
method when compared to alternatives such as LMNN.
\section{Explaining predictions using local metrics}
The metrics learned by our proposed method have two important characteristics:
\begin{description}
    \item[Diagonality]
    Each learned weight directly corresponds to a feature in the input space.
    If the input space itself is interpretable,
    so are our learned metrics.
    \item[Locality]
    For each point in our training set we find a local metric.
    If the learned metrics are interpretable,
    they tell us about how individual samples contribute to a prediction.
\end{description}
Consider a point \(x\) with a local metric that has a very small weight for some feature \(a\)
and a very large weight for another feature \(b\).
A second point \(y\) can be extremely different from \(x\) with respect to feature \(a\)
and still be close to \(x\) in terms of the local metric,
if the two points are similar with respect to feature \(b\).

When we use LANN to predict a label for an input,
the local metrics at the \(k\) nearest neighbors responsible for the prediction give us a distribution over the feature relevances involved in the prediction.
These distributions can be used directly for any given prediction
or aggregated to gain insight on different levels of detail;
the specifics depend on the down-stream task.

To demonstrate the usefulness and applicability of our learned local metrics in this sense,
we analyze the quality of embeddings induced by them.
\section{Interpreting learned metrics via low-dimensional
embeddings}\label{section:embeddings}
Reducing the dimensionality of data, as
preprocessing or for visualization, has a long history.
\Textcite{Pearson1901LinesPlanes} introduced the linear \emph{Principal
Component Analysis} in 1901, to determine -- and project onto -- the most
important directions in the original feature space; in 1969,
\textcite{Sammon1969NonlinearMapping} proposed the non-linear \emph{Sammon
Mapping} to find low-dimensional representations with locally faithful
differences. More recently, \emph{Uniform Manifold Approximation and Projection}
(UMAP)~\autocite{McInnes2018UMAP} has emerged as a versatile technique to
preserve topology during dimensionality reduction, arguably replacing
\emph{t-Distributed Stochastic Neighbor Embedding}
(t-SNE)~\autocite{Maaten2008tSNE} as state of the art in dimensionality
reduction.

The above-mentioned GMLVQ and LMNN,
each learning a global metric, also function to reduce dimensionality,
when the learned metric is used to project the data.
Both algorithms learn metrics to aid classification,
and so their embeddings are naturally discriminative,
meaning that even with their reduced dimensionality they allow to discriminate between classes.

In addition to embedding data in (low-dimensional) space, when a dimensionality
reduction algorithm (such as UMAP) uses differences between points, we can use
learned metrics during embedding to not only observe the data, but also the
respective metrics. Therefore, we can directly use our local metrics learned as
described in \cref{section:lann} to obtain an embedding that, if the metrics
behave properly, should be discriminative, similarly to LDA and GMLVQ.

Because LGMLVQ learns one metric per prototype, it is not immediately clear
which metric to use when determining the distance between any two given points.
We overcome this by first mapping each point \(\vec x\) onto its distance to
every one of the \(n\) prototypes \(\vec p_1, \dots, \vec p_n\):

\begin{equation}
    \vec x \mapsto (d_i(\vec x, \vec p_i))_{i = 1, \dots, n}^\transpsymb \in \R^n ,
\end{equation}

where \(d_i\) is the learned metric of prototype \(\vec p_i\). We subsequently
find a low-dimensional embedding of this \(n\)-dimensional space using UMAP.
\section{Experiments}
\subsection{Classification}\label{section:classification}
\begin{table}
    \centering
    \caption{Accuracies (cross validation averages) for all algorithms
    and datasets considered in our experiments.
    All but the first constitute real-world data.}
    \begin{adjustbox}{max width=\linewidth}
    \begin{tabular}{lrrrr}
        \toprule
        Dataset             & kNN               & LMNN              & LGMLVQ            & LANN\\
        \midrule
        Art. Classification & 0.95 $\pm$ 0.0029 & 0.97 $\pm$ 0.0042 & 0.99 $\pm$ 0.0017 & 0.99 $\pm$ 0.0018\\
        Adrenal             & 0.82 $\pm$ 0.0293 & 0.81 $\pm$ 0.0550 & 0.77 $\pm$ 0.0391 & 0.88 $\pm$ 0.0171\\
        Breast Cancer       & 0.95 $\pm$ 0.0079 & 0.95 $\pm$ 0.0113 & 0.92 $\pm$ 0.0155 & 0.94 $\pm$ 0.0077\\
        Digits              & 0.94 $\pm$ 0.0075 & 0.96 $\pm$ 0.0050 & 0.87 $\pm$ 0.0168 & 0.96 $\pm$ 0.0054\\
        Gamma Telescope     & 0.82 $\pm$ 0.0024 & 0.82 $\pm$ 0.0029 & 0.84 $\pm$ 0.0036 & 0.83 $\pm$ 0.0027\\
        Image Segmentation  & 0.93 $\pm$ 0.0039 & 0.95 $\pm$ 0.0064 & 0.94 $\pm$ 0.0051 & 0.95 $\pm$ 0.0041\\
        Ionosphere          & 0.77 $\pm$ 0.0468 & 0.79 $\pm$ 0.0353 & 0.76 $\pm$ 0.0375 & 0.90 $\pm$ 0.0306\\
        Iris                & 0.93 $\pm$ 0.0340 & 0.95 $\pm$ 0.0152 & 0.93 $\pm$ 0.0298 & 0.96 $\pm$ 0.0120\\
        Letter Recognition  & 0.88 $\pm$ 0.0019 & 0.91 $\pm$ 0.0027 & 0.87 $\pm$ 0.0072 & 0.91 $\pm$ 0.0025\\
        Outdoor Objects     & 0.80 $\pm$ 0.0070 & 0.83 $\pm$ 0.0084 & 0.83 $\pm$ 0.0117 & 0.87 $\pm$ 0.0085\\
        Pen Digits          & 0.98 $\pm$ 0.0012 & 0.99 $\pm$ 0.0008 & 0.99 $\pm$ 0.0016 & 0.99 $\pm$ 0.0011\\
        Robot Navigation    & 0.79 $\pm$ 0.0094 & 0.80 $\pm$ 0.0081 & 0.79 $\pm$ 0.0067 & 0.83 $\pm$ 0.0088\\
        USPS                & 0.94 $\pm$ 0.0025 & 0.95 $\pm$ 0.0023 & 0.95 $\pm$ 0.0027 & 0.95 $\pm$ 0.0024\\
        Wine                & 0.95 $\pm$ 0.0160 & 0.96 $\pm$ 0.0164 & 0.63 $\pm$ 0.0880 & 0.96 $\pm$ 0.0153\\
        \bottomrule
    \end{tabular}
    \end{adjustbox}
    \label{table:scores}
\end{table}

We have implemented our proposed algorithm (henceforth referred to as LANN) in
Python~3.7 within the
scikit-learn\footnote{\url{https://scikit-learn.org/}}~\autocite{scikit-learn}
framework, restricting the metrics to diagonal matrices as discussed above. We
compare its performance against a standard kNN classifier (as provided by
scikit-learn), against LMNN with a global adaptive metric (via the
implementation PyLMNN\footnote{\url{https://github.com/johny-c/pylmnn}} by John
Chiotellis) -- we keep $k = 5$ fixed for all three algorithms to facilitate
comparability -- and against Localized Generalized Matrix Learning Vector
Quantization (LGMLVQ -- using the open
implementation\footnote{\url{https://github.com/MrNuggelz/sklearn-lvq}} for
scikit-learn). Each algorithm is fitted and evaluated on a number of datasets:
\begin{description}[nosep]
    \item[Artificial Classification]
    An artificial dataset provided by scikit-learn that contains strongly relevant features,
    weakly relevant features, as well as redundant features.
    We sample \num{2000}~data points according to the default parameters,
    which results in \num{2}~classes,
    \num{20}~features, of which \num{2} are strongly relevant,
    and \num{2} are weakly relevant.
    \item[Adrenal]\autocite{Biehl2012MatrixRelevance}
    Results from an analysis of adrenal gland metabolomics.
    The dataset contains \num{147}~data points in \num{2}~classes (adrenocortical carcinoma and adenoma),
    described by \num{32}~features that relate to the underlying metabolic processes.
    \item[Wisconsin Breast Cancer]
    Classic dataset of \num{569}~data points in \num{2}~classes (benign and malignant)
    described by \num{30}~features that relate to the properties of cells visible under a microscope.
    \item[Digits]\autocite{Alpaydin1998Cascading}
    \num{1797}~images of \num{8}~by \num{8}~pixels that contain handwritten digits (\num{10}~classes).
    \item[Gamma Telescope]\autocite{Heck1998Corsika}
    Registration of high-energy gamma particles in a telescope.
    The dataset contains \num{19020}~samples with \num{11}~features in two classes (signal and background).
    \item[Image Segmentation]\autocite{Dua2017UCI}
    In this dataset,
    \num{2306}~data points fall into \num{7}~classes
    and are described by \num{16}~features that encode several attributes of image regions.
    We leave out three near-constant features,
    as suggested by \textcite{Schneider2009AdaptiveRelevanceMatrices}.
    \item[Ionosphere]\autocite{Sigillito1989Ionosphere}
    Electrons in the ionosphere recorded by a high-frequency radio antenna array.
    The binary dataset contains \num{351}~samples with \num{34}~features.
    \item[Iris]\autocite{Fisher1936MultipleMeasurements}
    Classic dataset of \num{150}~samples in \num{3}~classes that are three different types of the plant Iris.
    The \num{4}~features are sepal and petal length and width, respectively.
    \item[Letter Recognition]\autocite{Frey1991LetterRecognition}
    Based on black-and-white images of capital letters (corresponding to \num{26}~classes),
    this dataset consists of \num{20000}~samples of \num{16}~hand-crafted features.
    \item[Outdoor Objects]\autocite{Losing2016KNN}
    Here, \num{4000}~data points correspond to images that belong to one of \num{40}~classes,
    depending on objects visible in the images.
    Its \num{21}~features constitute normalized color histograms.
    \item[Pen Digits]\autocite{Alimoglu1996PenDigits}
    Recognition of handwritten digits,
    based on readings from a stylus and a pressure-sensitive tablet.
    The dataset consists of \num{10992}~samples of \num{16}~features in \num{10}~classes.
    \item[Robot Navigation]\autocite{Freire2009RobotNavigation}
    Ultrasound sensor readings obtained by a robot during navigation.
    The \num{5456}~samples are represented by \num{24}~features and the \num{4}~classes correspond to directional movement instructions.
    \item[USPS]\autocite{Hull1994USPS}
    Another dataset for handwritten digit recognition that contains \num{16}~by \num{16}~pixel images.
    The data was originally obtained in cooperation with the US Postal Service.
    \item[Wine]
    Classic dataset with \num{3}~classes (types of wine), \num{13}~features, and \num{178}~samples.
\end{description}

For each algorithm and dataset we perform a 10-fold, stratified, randomly
shuffled cross validation and include a z-score transformation as the only
preprocessing step. We report the averaged accuracies together with their
standard deviations in \cref{table:scores}. LANN obtains an improvement as
compared to LMNN in four out of five cases, yielding a smaller variation in all
cases. Interestingly, local metric learning seams particularly profitable for
the outdoor objects data, a setting with a large number of classes and
comparably high degree of noise. 

\begin{figure}
    \centering
    \includegraphics[width=\linewidth]{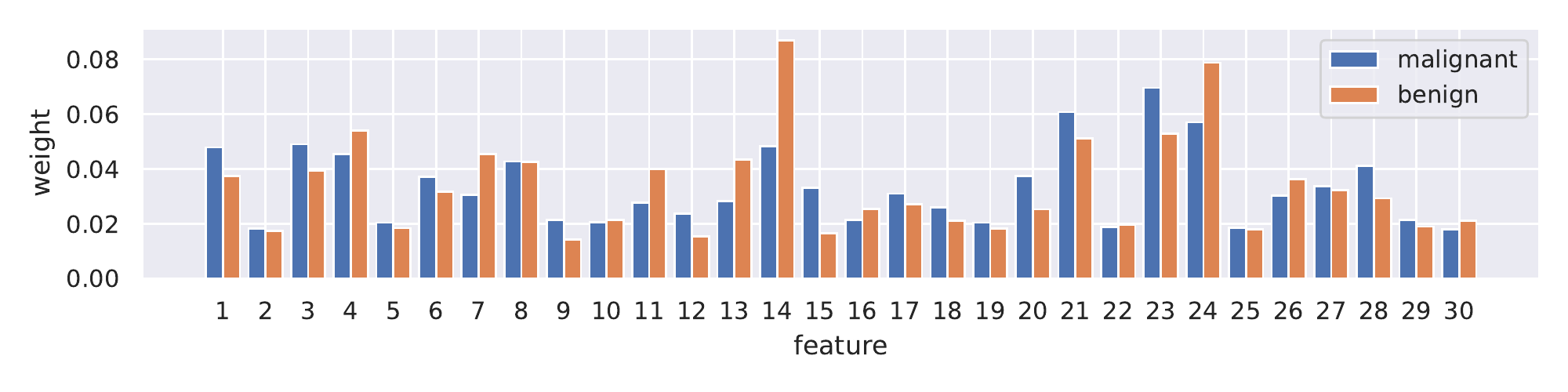}
    \caption{Aggregated per-class relevances for the Wisconsin Breast Cancer
    dataset, as determined by our proposed algorithm. The two different colors
    indicate the two classes \emph{benign} and \emph{malignant}.}
    \label{plot:wisconsin-fingerprints}
\end{figure}

LANN yields an indication of relevance for each feature with respect to each
individual data point. We can use these to develop a \emph{local} understanding
of feature relevance. For the Wisconsin Breast Cancer dataset, we aggregate
these relevances class-wise. Our findings, presented in
\cref{plot:wisconsin-fingerprints}, align with those previously discovered and
discussed in the literature~\autocite{gopfert2018FeatureRelevance}. In
particular, it becomes apparent that different averages result for the two
classes.
\subsection{Embeddings}
\begin{figure}
    \centering
    \includegraphics[width=\linewidth]{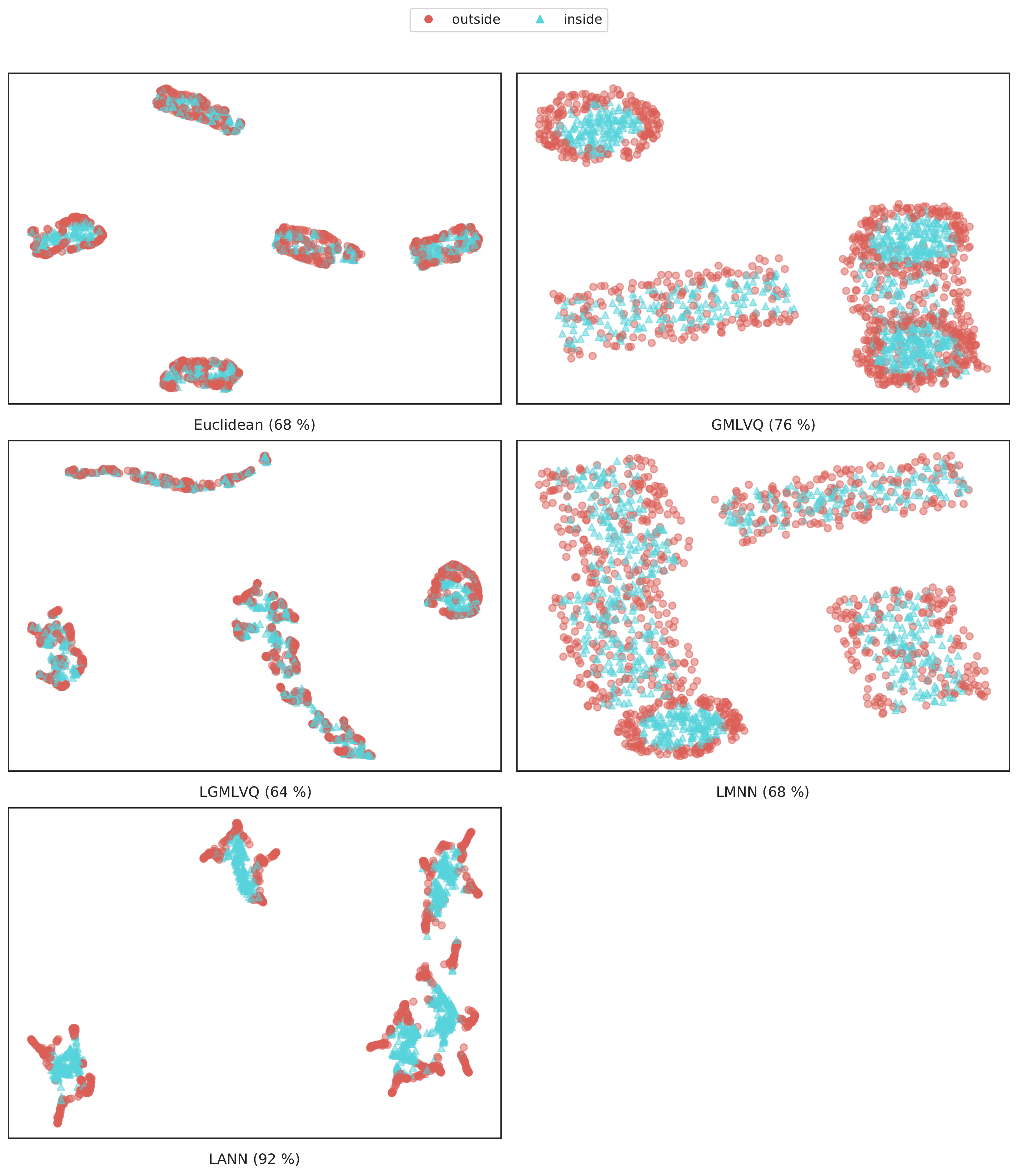}
    \caption{Embeddings of an artificial data set (Licorice), which consists of
             five distinct cylinders that in turn contain points labeled
             according to whether they are inside the cylinder or on its
             outside. Below each plot, we indicate the metric or the algorithm
             (that produces it) used to obtain the embedding, as well as the
             accuracy when classifying based on the embedding. Class labels are
             indicated by color and shape. Notably, the local metrics of LANN
             ensure that “inside” points are placed close to one another with
             “outside” points spread around them, irrespective of the underlying
             cylinder's orientation -- at the same time, the global metrics of
             GMLVQ and LMNN struggle to cope with these different orientations.}
    \label{embedding:licorice}
\end{figure}

\begin{figure}
    \centering
    \includegraphics[width=\linewidth]{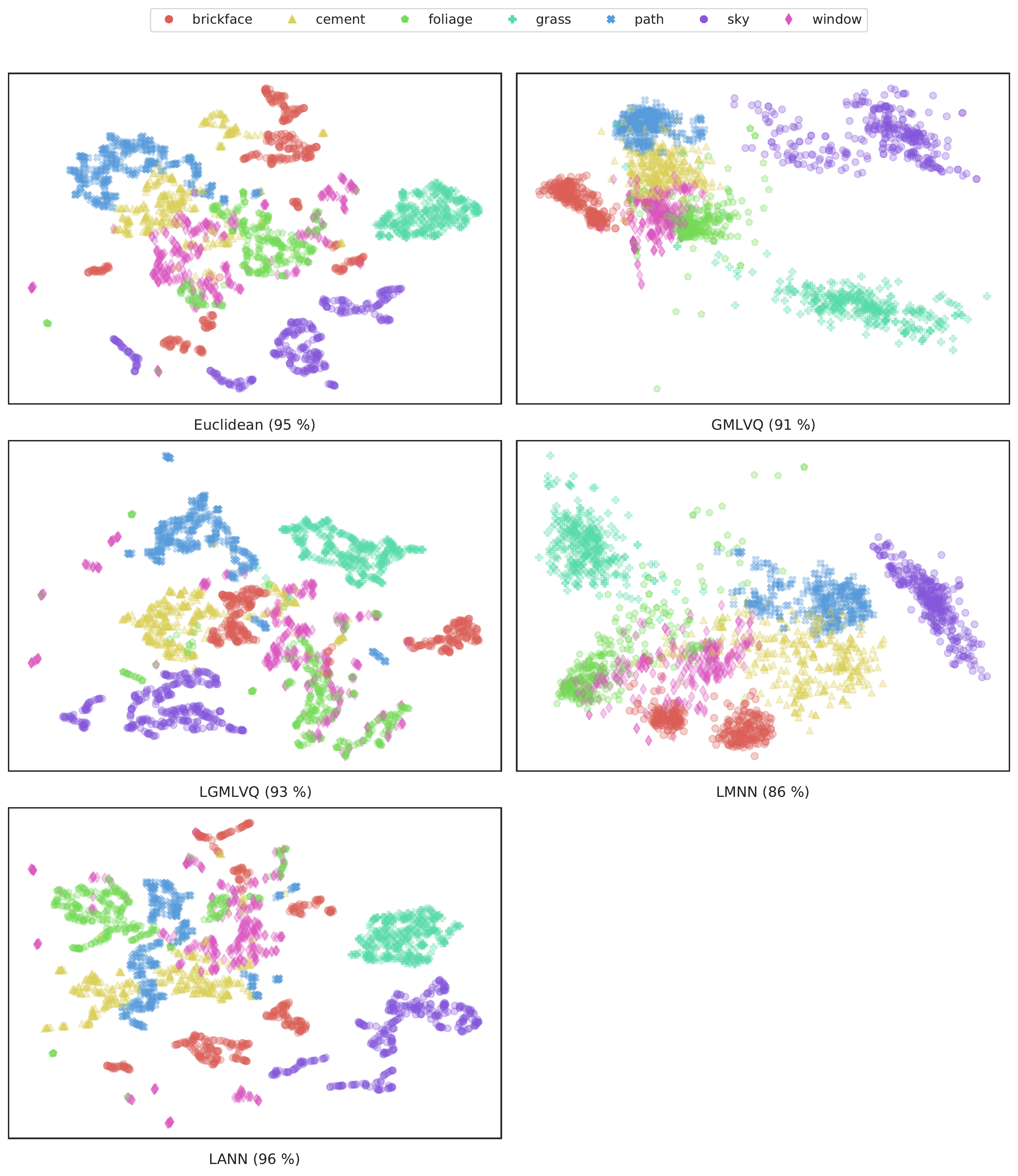}
    \caption{Embeddings of the real-world data set \emph{Image Segmentation}.
             Below each plot, we indicate the metric or the algorithm (that
             produces it) used to obtain the embedding, as well as the accuracy 
             when classifying based on the embedding. Class labels are indicated
             by color and shape.}
    \label{embedding:image-segmentation}
\end{figure}

To asses and visualize the nature of and relation between the local metrics LANN
finds, we compute embeddings for two data sets: one artificial data set where we
know that locality is crucial, and \emph{Image Segmentation} (see
\cref{section:classification}) as a real-world data set. We present the results
in \cref{embedding:licorice,embedding:image-segmentation}. The artificial data
set, which we dub \emph{Licorice}, consists of several distinct cylinders of
varying orientation that in turn contain points labeled according to whether
they are located inside the cylinder or on its outside.

For each data set, we first compute an embedding via UMAP using the Euclidean
metric. We then train a set of algorithms on the entire data set; the global
low-rank metrics learned by GMLVQ and LMNN directly lead to embeddings; LANN
yields a local metric for each point, such that we can use pairwise distances as
input for UMAP; and for LGMLVQ we compute a proxy embedding as described in
\cref{section:embeddings}, which is in turn embedded by UMAP.


To quantify the quality of the produced embeddings with respect to their
discriminative power, we reclassify each point by its nearest neighbors in said
embedding, and indicate the proportion of correctly classified points below the
respective plots. Note that these numbers are not to be taken as deciding scores
with regards to the efficacy of the respective algorithms; they merely aid in
interpreting the validity of the embeddings. Because UMAP preserves local
neighborhoods, its embeddings naturally lend itself well to nearest neighbor
classification.

All algorithms result in viable low-dimensional embeddings of the original data.
As expected, and in line with our findings in \cref{section:classification},
local metrics do improve the discriminative power when used for embedding.
Especially in our artificial data set, where cylinders are present in different
orientations, the ability to adapt to local properties is crucial. Furthermore,
the results on \emph{Image Segmentation} underline the usefulness of LANN,
enabling a better distinction even between classes “foliage” and “window”, which
appears to be particularly challenging.

\section{Conclusions}
We have proposed a metric learning scheme which assigns a separate relevance
weighting vector to every data point of a kNN classifier, leading to different
local relevances of the decision function. Even restricted to local diagonal
matrices, the technology is as good as or surpasses popular metric learning
schemes such as LNMM. More importantly, the method provides a local explanation
of a specific decision of the model given an input $\vec x$ rather than a global
metric, and it enables online update rules in the form of a stochastic gradient.
We have demonstrated the quality and applicability of the learned local metrics
via low-dimensional embeddings obtained through state of the art dimensionality
reduction.

\subsection{Limitations \& Future Work}
It is subject to future work to integrate this scheme into kNN
methods for streaming data and to investigate the suitability to build a reject
option based on this representation, as investigated in
\cite{samknn,Goepfert2018Mitigating} for the standard Euclidean metric.

Due to the local metrics,
common optimizations for nearest neighbor computations are not readily applicable to LANN,
so we cannot currently recommend it for big data computations.
However, because during our proposed iterative updates are local,
we see a number of promising directions to optimize computations.

How well explanations perform is difficult to quantify,
because what constitutes a good explanation depends on concrete applications,
and because ground truth is not usually available.
In our setting, this issue is compounded by a lack of ground truth for local relevances.
Our proposed method can aid in exploratory data analysis accompanied by interactive explanations for predictions,
and we are looking forward to releasing a framework for this purpose based on LANN.

%% file: acknowledgements.tex
\section*{Acknowledgements}
We gratefully acknowledge support by Honda Research Institute Europe.